\documentclass[authoryear,12pt]{elsarticle}

\usepackage{amssymb}
\usepackage{graphicx}

\begin{document}

\begin{frontmatter}

\title{Wood Surface Inspection Using Structural and Conditional Statistical Features}
\author{Cem \"{U}nsalan}
\address{Department of Electrical and Electronics Engineering,\\
Yeditepe University, \.{I}stanbul 34755, Turkey.\\
unsalan@yeditepe.edu.tr\\}

\begin{abstract}
Surface quality is an extremely important issue for wood products in the market. Although quality inspection can be made by a human expert while manufacturing, this operation is prone to errors. One possible solution may be using standard machine vision techniques to automatically detect defects on wood surfaces. Due to the random texture on wood surfaces, this solution is also not possible most of the times. Therefore, more advanced and novel machine vision techniques are needed to automatically inspect wood surfaces. In this study, we propose such a solution based on support region extraction from the gradient magnitude and the Laplacian of Gaussian response of the wood surface image. We introduce novel structural and conditional statistical features using these support regions. Then, we classify different defect types on wood surfaces using our novel features. We tested our automated wood surface inspection system on a large data set and obtained very promising results.
\end{abstract}


\end{frontmatter}

\section{Introduction}

Consumers use wood products extensively in everyday life. For most wood products (such as office furniture) surface quality is utmost important. Therefore, during manufacturing experts have been grading wood surfaces based on their defects. However, a simple eye examination is not sufficient to inspect and grade wood surfaces with their unpredictable texture and color variations. Therefore, there is a strong demand to automate the wood surface quality inspection process. In the introduction to the special issue of \emph{Computers and Electronics in Agriculture}, \cite{Kline2} clearly explain why automated defect detection systems are needed for wood products. They emphasize that, with an automated defect detection system, low cost and high quality wood products can be obtained. Therefore, researchers focused on automated machine vision systems (using different sensors) to solve this problem.

\cite{Kauppinen5} focused on color calibration on wood defect inspection. They used percentiles as statistical features and were able to eliminate $85 \%$ of the sound wood. Then, they applied segmentation by thresholding followed by connected components analysis to find the location, dimension, and shape features for each blob. \cite{Smolander1} considered self-organizing map (SOM) based feature extraction in wood surface inspection. They used Gray Level Co-occurrence Matrix (GLCM), Gabor, and percentile features and obtained a $85 \%$ recognition rate over 400 grayscale knot images. \cite{Kauppinen1} mentioned that, human visual inspection rarely achieves better than $70 \%$ performance in grading lumber. They also mentioned that, thresholding based defect detection methods (especially in extracting the structure) do not perform well. They tested percentiles, second order texture features such as local binary patterns, and statistical features calculated from GLCM. They obtained a $95\%$ defect detection performance with a $22.9\%$ false alarm rate. \cite{Forrer1} focused on image segmentation for defect detection on wood surfaces. They concluded that, a region based algorithm exhibited the best overall performance. \cite{Perez1} also mentioned that, image segmentation is the crucial step for wood surface inspection. They proposed a neuro-fuzzy color image segmentation method for this purpose. \cite{Kurdthongmee} in a recent study, used color information with a SOM network to classify rubberwood boards. Other researchers considered the problem from the systems viewpoint. Some of these methods can be found in the cited references \citep{Kline1,Vogrig1,Bond1,Espinoza1,Drayer1,Araman1,Kim1}.

Although these valuable studies focused on automated wood surface inspection, none of them has considered the combination of the structural (such as defect shape) and statistical information on the wood surface. Fusion of structural and statistical information may drastically improve the defect detection performance. Therefore, in this study we introduce novel structural and conditional statistical features (using the fusion of structural and statistical information) to automatically detect and classify defects on wood surface images. We explain our wood surface defect detection method in four steps. In Section~\ref{sec:support}, we explain how to extract support regions from wood surface images. Then, in Section~\ref{sec:feature}, we propose novel structural and conditional statistical features using these support regions. In Section~\ref{sec:experimental}, we perform defect detection and classification tests using our novel features. Finally, we provide the strengths and weaknesses of our wood surface inspection method in Section~\ref{sec:final}.

\section{Support Regions}\label{sec:support}

A high intensity change in a wood surface image either indicates a natural texture or a defect. Therefore, the intensity change may be a valuable cue to detect defects from the wood surface image. One possible way to locate high intensity changes in images is using edge detection algorithms. In edge detection, each intensity profile is modeled by a specific function. To detect changes on a specific intensity profile, optimal filters (in general, either a gradient or a Laplacian filter) are introduced. In applying gradient filters, a high gradient magnitude indicates an intensity change. As for the Laplacian filter
response, zero crossings indicate an intensity change. After the filtering operation, pixels possibly belonging to edge locations are labeled based on various algorithms \citep{Boyle1}. As a result, edge pixels in the image are obtained.

Although the above approach is standard for computer vision, there are different approaches to extract the edge information in the image. In a previous study, we used gradient magnitude based support regions (GMSR) to classify land use in Ikonos satellite imagery \citep{Unsalan4}. In this study, we follow a similar strategy in extracting structural features to detect defects in wood surface images. Therefore, we first use GMSR. Then, we introduce a novel support region extraction method using the Laplacian of Gaussian filtering of the image. Since defects on wood surfaces have different characteristics, each can be detected by a different support region. Therefore, gradient magnitude and Laplacian of Gaussian response based support regions will be of use at the same time. To make the paper more complete, we briefly overview the GMSR next. Then, we explore our novel support region extraction method based on Laplacian of Gaussian filtering.

\subsection{Gradient Magnitude Based Support Region (GMSR) Extraction}

We benefit from the GMSR in extracting structural features and conditioning statistical features from wood surface images. For completeness, we summarize how to extract GMSR next. To obtain GMSR from an image, we first obtain optimally smoothed gradient vectors using two dimensional Gaussian filters:

\begin{equation} \label{eq:gradientx}
g_x(x,y) = \frac {-x} {2\pi\tau_g^4} \exp(-\frac {x^2+y^2} {2\tau_g^2})
\end{equation}

\begin{equation} \label{eq:gradienty}
g_y(x,y) = \frac {-y} {2\pi\tau_g^4} \exp(-\frac {x^2+y^2} {2\tau_g^2})
\end{equation}

\noindent where $\tau_g$ is the scale parameter for both filters. This scale is directly related to the minimum dimension of the defect to be detected. We tested and reported the effect of different scale values on detection performance in Appendix. Finally, we decide that $\tau_g=2$ is suitable for our application.

Let $I(x,y)$ be the grayscale image, the horizontal component of the smoothed gradient is $ G_x(x,y)= I(x,y)*g_x(x,y)$ where $*$ is the convolution operation. Similarly, the vertical component of the smoothed gradient is $G_y(x,y)= I(x,y)*g_y(x,y)$. The gradient magnitude at $(x,y)$ is $M(x,y) = \sqrt{ G_x^2(x,y)+ G^2_y(x,y)}$ where square and square root values are calculated on a pixel basis.

To extract support regions, we first apply hysteresis thresholding to $M(x,y)$. In this method, two threshold values are selected as high and low. First, we label $M(x,y)$ pixels having values greater than the high threshold. Then, $M(x,y)$ pixels neighboring these labeled pixels and having values greater than the low threshold are also labeled. All these labeled pixels form the candidate support regions as a binary image. Detailed information on hysteresis thresholding can be found in \citep{Boyle1}. We experimentally observed that setting the high threshold value to $0.2$ and low threshold value to $0.15$ works fairly well for the normalized (wrt. the maximum value) gradient magnitudes. It is important to note here that, our wood surface images are taken in a controlled environment. Therefore, constant threshold values can be used without any problem.

We next group pixels in the thresholded binary image based on their spatial neighborhood. In grouping, we benefit from binary image labeling \citep{Boyle1}. Each pixel group forms a separate gradient magnitude based support region (GMSR). The final binary image containing support regions are called as $SR(x,y)$. The magnitude calculations and the following operations are all rotation invariant \citep{Kak1}. Therefore, our gradient magnitude based support region extraction is also rotation invariant.

\subsection{Laplacian of Gaussian Based Support Region (LGSR) Extraction}

Similar to the gradient magnitude (based on first order partial derivatives), the Laplacian of Gaussian filtering (based on second order partial derivatives) of an image can be used to locate intensity changes. \cite{Hildreth1} introduced the Laplacian of Gaussian (LoG) filter to extract edges. In this method, the Laplacian operator performs the partial second order derivative calculation and the Gaussian filter smooths the image. By the property of convolution, both smoothing and Laplacian operations can be represented as the LoG filter:

\begin{equation} \label{eq:logfilter}
f(x,y) = \frac {1} {\pi\tau_l^4} \left[\frac {x^2+y^2} {2\tau_l^2} -1 \right] \exp(-\frac {x^2+y^2} {2\tau_l^2})
\end{equation}

\noindent where $\tau_l$ is the scale parameter of the filter. As in the previous section, the scale parameter is directly related to the minimum defect size to be detected. After extensive testing, we set $\tau_l=1$ in this study. Hence, we have a $23 \times 23$ mask for LoG filtering.

Marr and Hildreth extracted edge locations in the image by labeling zero crossings of the LoG filter response. Several authors also mention the usage of the LoG filter for blob detection in images \citep{Ahuja1}. For our application, we define the LoG based support regions in the same way as blob detection. We first convolve the image with the LoG filter as $M(x,y) = I(x,y)*f(x,y)$. As if detecting blobs in the image, we apply thresholding on $M(x,y)$. Then, we group pixels in the thresholded image in the same line with the GMSR extraction. Finally, we obtain the separate LoG based support regions. We call them as LGSR. To note here, the LoG filter is rotation invariant due to its circular symmetry. This leads to rotation invariant support region extraction.

\subsection{Examples}

\begin{figure}[htbp]
\centering
\includegraphics[width=4.5 in]{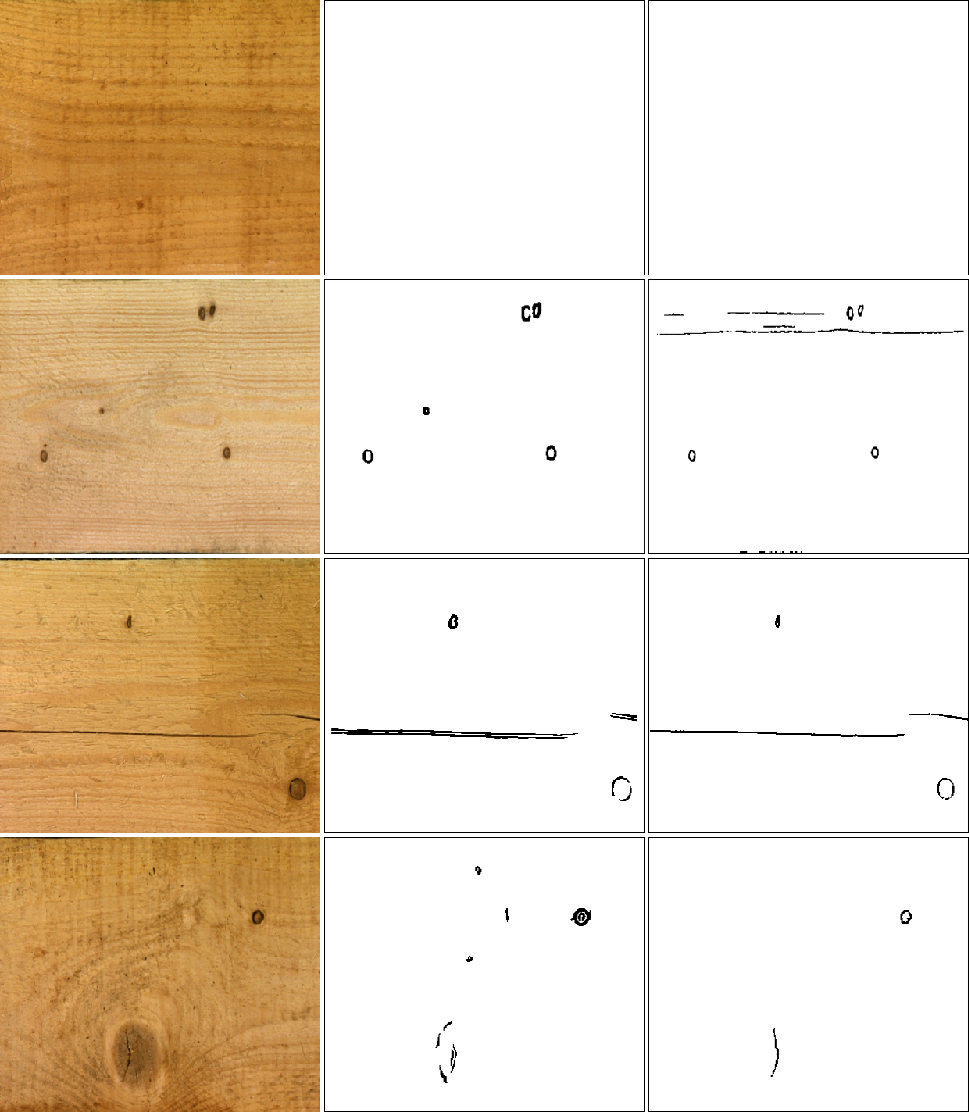}
\caption{Wood surface examples and support regions extracted from them. First column: wood surface images (non-defective wood surface, wood surface with knots, wood surface with shakes and knots, wood surface with a sound and dry knot); second column: GMSR results; third column: LGSR results.} \label{fig:wood_types}
\end{figure}

We provide support region extraction examples on various wood surface images next. There are four different wood surfaces and support regions extracted from them in Fig.~\ref{fig:wood_types}. In the first row of the figure, there is a non-defective wood surface. In the second row, the wood surface has knots. In the third row, the wood surface has shakes and knots. In the fourth row, the wood surface has a sound and dry knot. The second column of Fig.~\ref{fig:wood_types} corresponds to the GMSR extracted from each wood surface. Similarly, the third column corresponds to the LGSR extracted.

As can be seen, both support region extraction methods performed well (no support regions are extracted) on the non-defective wood surface image in the first row. For the defective wood surface image given in the second row, all knots have been detected successfully by the GMSR. There are some false alarms for the LGSR. Knots and shakes are detected correctly by both support region extraction methods in the third row. The dry knot is detected by both support region extraction methods in the fourth row. However, both support region extraction methods can not extract the sound knot reliably. Besides, GMSR gives some false alarms for this image.

\subsection{Support Region Representation}

To capture the structure information, we extract curves corresponding to support regions. To do so, we first obtain the outer boundary of each support region. The boundary can be represented by a complex periodic function $b(t)=b_x(t)+jb_y(t)$; $t=0,...,T-1$ and $j=\sqrt{-1}$. In our previous study, we used a second order Fourier series expansion to extract straight lines from support regions \citep{Unsalan3}. Since we are dealing with curves here, we use the fourth order Fourier series expansion as

\begin{equation} \label{eq:fsf2}
\hat{b}(t)=\sum_{n=0}^{4}B_n e^{j\frac {2\pi nt} {T}}
\end{equation}

\noindent where

\begin{equation} \label{eq:fsf4}
B_n=\frac {1} {T} \sum_{t=0}^{T-1}b(t) e^{-j\frac {2\pi nt} {T}}
\end{equation}

\noindent To note here, a higher order Fourier series expansion can also be used. However, this may complicate the curve extraction process explained next.

The curvature of the boundary representation in Eqn.~\ref{eq:fsf2} leads to curve extraction from the corresponding support region. The curvature is a differential geometric entity giving a measure of how rapidly the curve deviates from the tangent line \citep{Docarmo1}. We find the curvature of $\hat{b}(t)=\hat{b}_x(t)+j\hat{b}_y(t)$, as

\begin{equation} \label{eq:curvature}
K(t)= \frac { \frac {d\hat{b}_x(t)} {dt} \frac {d^2\hat{b}_y(t)} {dt^2} - \frac {d\hat{b}_y(t)} {dt} \frac {d^2\hat{b}_x(t)}
{dt^2} } {\left( (\frac {d\hat{b}_x(t)} {dt})^2 + (\frac {d\hat{b}_y(t)} {dt})^2 \right)^{3/2}}
\end{equation}

Extrema of the curvature correspond to the endpoints of the curve (on the boundary) to be extracted. Assume that we obtain two extrema as $a_1,a_2$, such that each corresponds to an endpoint of the curve. We divide the approximated boundary into two parts $\hat{b}_{xu}(t)+j\hat{b}_{yu}(t)$ (the upper part of the boundary) and $\hat{b}_{xl}(t)+j\hat{b}_{yl}(t)$ (the lower part of the boundary) based on these extrema. Without loss of generality, let's assume that these two parts have the same parameter such that $t=[0,a]$. They also have the same increment direction. Our final curve $s(t)$ is

\begin{equation} \label{eq:csffinal}
s(t)=\frac{1} {2} \left[\left( \hat{b}_{xu}(t)+ \hat{b}_{xl}(t) \right) +j \left( \hat{b}_{yu}(t)+ \hat{b}_{yl}(t)
\right)\right]
\end{equation}

\begin{equation} \label{eq:csffinal2}
s(t)=s_x(t)+js_y(t)
\end{equation}

\noindent where $t=[0,a]$.

\begin{figure}[htbp]
\centering
\includegraphics[width=4.5 in]{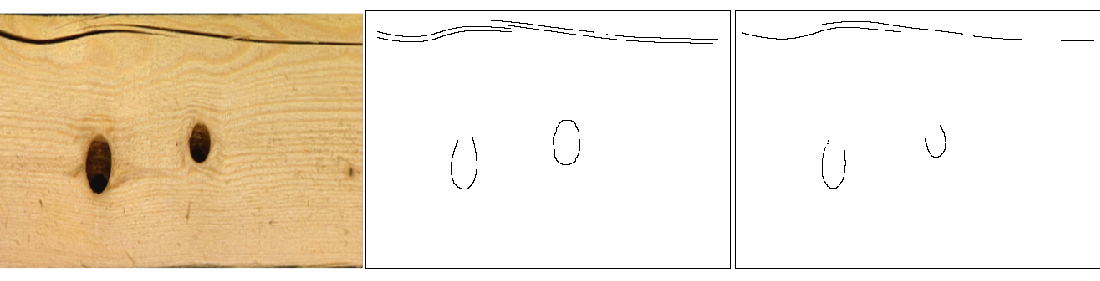}
\caption{A sample wood surface image, corresponding GMSR and LGSR based curves (from left to right).} \label{fig:shakeknotcurves}
\end{figure}

To demonstrate the curve extraction method, we pick a defective wood surface image as in Fig.~\ref{fig:shakeknotcurves}. This wood surface has two knots and a shake. We extract the GMSR and LGSR based curves as explained above. As can be seen, curves are successfully extracted from both support region extraction methods.

\section{Feature Extraction}\label{sec:feature}

In order to detect and classify defects in wood surfaces, we extract features using support regions and the statistical information. Therefore, we consider three sets of feature extraction methods. The first set consists of statistical features. Although statistical features are introduced earlier, we report them here for completeness and comparison purposes. The second set consists of novel structural features obtained from support regions. The third set consists of novel conditional statistical features.

\subsection{Statistical Features}

Our first set of features is based on the sample statistics calculated over a given image window as: the sample mean ($\mu$), the sample standard deviation ($\sigma$), the sample skewness ($\gamma_1$), the sample kurtosis ($\gamma_2$), the median ($\mu_{\tilde{m}}$), and percentiles. These statistical definitions can be found in \cite{Devore}. Although the first five statistics are used frequently, percentiles may not be known by all readers. Devore, defines the percentile as

\begin{equation}\label{percentile}
p=\int^{\eta(p)}_{-\infty}f(\alpha)d\alpha
\end{equation}

\noindent where $f(\alpha)$ is the sample probability density function. The percentile $\eta(p)$ is the number on the measurement scale, such that the area under the density curve to the left of $\eta(p)$ is the probability $p$. Following \cite{Kauppinen1}, we tested five different percentile values ($0.02\%$, $0.2\%$, $10\%$, $60\%$, and $90\%$). We label them as $p_{v \in \{0.02, 0.2, 10, 60, 90\}}$.

\subsection{Structural Features}

We base our second set of features on the geometric properties of curves extracted from support regions. These features are calculated for each support region separately. Therefore, we do not need a constant window size, which causes problems in most applications. However, in order to compare these structural features with the previous statistical ones, we apply the following strategy. Each feature is calculated for all support regions in the image window. Then, the window is represented by the value of the dominant feature (having more pixels in its support region) within its domain.

Since knots and shakes occur more often in wood surfaces, we design our structural features to detect them specifically. Therefore, all the structural features mentioned in this section try to grab the same information in different ways. However, these features are not redundant. Their classification performances also emphasize this non-redundancy experimentally. In listed form, the structural features based on the geometric properties of curves are: Normalized endpoint distance ($d_n$), compactness ($c$), median of curvature ($k$), and ellipse axes ratio ($e$). We define these features in detail next.

\subsubsection{Normalized Endpoint Distance ($d_n$)}

This feature depends on the ratio of the distance between endpoints of the curve to its arc length. Based on $s(t)$, $t=[0,a]$ in Eqn.~\ref{eq:csffinal2}, the normalized endpoint distance is

\begin{equation} \label{eq:endpoint_distance}
d_n=\frac{ |s(0)-s(a)|} {\int_{t=0}^{t=a} \sqrt{(\frac {ds_x(t)} {dt})^2+(\frac{ds_y(t)} {dt})^2)} dt}
\end{equation}

\noindent $d_n$ is scaled between zero (corresponding to a closed shape, such as a knot) and one (corresponding to an open and elongated shape, such as a shake). This feature therefore can be used to discriminate elongated and closed shapes.

\subsubsection{Compactness ($c$)}

The second structural feature is based on the compactness of a support region. Compactness is defined as the ratio of a region's area to the square of its perimeter, normalized by $4\pi$ \citep{Boyle1}. For the GMSR, we can define the compactness as

\begin{equation} \label{eq:compactness}
c=\frac{ \sum_{x=0}^{N-1}\sum_{y=0}^{M-1} SR(x,y)} {4\pi T}
\end{equation}

\noindent where $T$ is the perimeter and $SR(x,y)$ is the binary image for the support region. Hence, the most compact shape (circle of any size) has a value one. Other shapes will have values greater than one. Most defects have relatively small areas and a smooth boundary (as extracted by our support regions) compared to the texture on the wood surface. We can extract this information using compactness.

\subsubsection{Median of Curvature ($k$)}

The third structural feature is based on the curvature given in Eqn.~\ref{eq:curvature}. To capture the curvature information (as a feature) we calculate its median, $k$. For elliptical curves, the curvature has very high values. On the contrary, for straight lines, the curvature has low values. For other curves, the median of the curvature has values between these two extrema. Similar to the previous features, we plan to discriminate support regions corresponding to defective and non-defective regions using this information.

\subsubsection{Ellipse Axes Ratio Feature ($e$)}

Our last structural feature is based on ellipse fitting to a support region \citep{Unsalan7}. As the ellipse is fit to the support region, we define the ratio of its major and minor axes lengths as a new feature

\begin{equation} \label{eq:ellipse_axes}
e=\frac{\lambda_1} {\lambda_2}
\end{equation}

\noindent where $\lambda_1$ is the length of the major axis; $\lambda_2$ is the length of the minor axis of the ellipse.  This feature gives a low value for circular curves and a high value for straight lines.

\subsection{Conditional Statistical Features}

Our last set of features combine the structural and statistical information. In statistical feature extraction, we did not take the structure information into account. In the structural feature extraction, we did not consider the texture information on the surface. Their combination may lead to more discriminative features. Combination is based on conditioning the statistical features over the structure extracted. In our previous studies, we followed a similar strategy \citep{Unsalan3,Unsalan7}. Here, we extend their usage to new features.

In extracting conditional statistical features, we assume that support regions act as a spatial filter emphasizing possible defective regions. Therefore, in calculating statistical features, only highly probable pixels (belonging to support regions) are considered. If we do not apply such filtering, all pixels in the image window are taken into account. This may decrease the discrimination performance of the feature. Therefore, we expect to improve the discrimination performance of the feature as we introduce a spatial condition.

We explain our conditional statistical feature calculation on conditional sample mean calculation. Let $I(x,y)$ be the image with size $N \times M$. The sample mean is

\begin{equation} \label{eq:sample_mean}
\mu=\frac{\sum_{x=0}^{N-1}\sum_{y=0}^{M-1} I(x,y)} {NM}
\end{equation}

\noindent If we use GMSR as a spatial condition, then the conditional sample mean becomes

\begin{equation} \label{eq:conditional_sample_mean}
\mu^G=\frac{\sum_{x=0}^{N-1}\sum_{y=0}^{M-1} I(x,y)SR(x,y)} {\sum_{x=0}^{N-1}\sum_{y=0}^{M-1} SR(x,y)}
\end{equation}

\noindent where $SR(x,y)$ is the GMSR representation for the image $I(x,y)$.

We denote the conditional sample mean as $\mu^G$ (with a superscript $G$) to indicate that, GMSR is used as a spatial condition. If we were to use the LGSR, then only $SR(x,y)$ should be taken accordingly. Then we would represent it as $\mu^L$ (with a superscript $L$), to indicate that the spatial condition is the LGSR. Other conditional statistical features are defined in the same way. Following this strategy, novel conditional statistical features we introduce in this study are: Conditional mean ($\mu^G$, $\mu^L$); conditional standard deviation ($\sigma^G$, $\sigma^L$); conditional skewness
($\gamma_1^G$, $\gamma_1^L$); conditional kurtosis ($\gamma_2^G$, $\gamma_2^L$); conditional median ($\mu_{\tilde{m}}^G$, $\mu_{\tilde{m}}^L$); and conditional percentiles ($p_{v}^G$, $p_{v}^L$).

\section{Experimental Results}\label{sec:experimental}

We test our features on detecting and classifying defects on a diverse and representative wood surface image set. Therefore, we use Bayes classifiers (in a hierarchical manner) starting with defect detection. This is followed by classification of sound knot, dry knot, and elongated defects (such as shakes). Since these defects are observed often, we pick them as our classes.

The wood surface image data set is obtained from Oulu University, Information Processing Laboratory. Defects on these images are also labeled (in a separate text file) by an expert. From this data set, we picked 744 wood surface images. Each image is approximately $500 \times 400$ pixels. As in previous studies, we formed $60 \times 60$ pixel subwindows to calculate our features. This window size is picked in accordance with the average knot size to be detected. By forming these subwindows, we obtained 24886 samples for classification. In all classification levels, we only provide the performances on ``test samples''
(discarding samples used for training).

In structural and conditional statistical feature calculations, we discard support regions smaller than 50 pixels assuming that they do not represent any structure reliably. The computation cost of each feature directly depends on the number of support regions present in the image window. As the number of support regions increase, computations for each feature will also increase. Therefore, eliminating ``small'' support regions also helps us to improve the feature extraction process. In order to show the effect of this elimination, we tested and tabulated different support region elimination sizes in Appendix.

We tested different color spaces in calculating features. We conclude that the ``red'' and ``green'' bands of the RGB color space, and the ``saturation'' and ``value'' bands of the HSV color space are better than the rest in accordance with previous studies. Although color images increase the computational cost in feature extraction, they hold valuable information. In the following sections, the color space used for each feature is given as a subscript within parenthesis as: $(r)$, $(g)$, $(s)$, and $(v)$ to represent the red, green, saturation, and value bands respectively.

At each classification level, we apply a global search on all features extracted from all color bands. Since each level needs a different type of information for discrimination, this search strategy is necessary. In tabulating results, we only provide the best performing subset for each feature set. Our feature sets are: statistical, GMSR based (both structural and conditional statistical), and LGSR based (again both structural and conditional statistical). We also combine all features as one set and obtain the best performing subset from it.

Besides the standard Bayes classifier, we also tested a Neyman-Pearson type decision represented by the Receiver Operation Characteristics (ROC) curves on probability of detection, $P_d$, versus probability of false alarm, $P_f$. To form ROC curves, we shift the Bayes decision boundary in the classifier and obtained $P_d$ vs. $P_f$ successively. Therefore, the possible user can select the performance range suitable to his or her application.

Besides the Bayes classifier, we tested different classifiers on detecting defects. We also tabulate the effect of different parameters on defect detection. Since these variations may give insight to a potential user, we tabulate them in Appendix. 

\subsection{Defect Detection}

The first level in classification is detecting all types of defects in wood surfaces. Therefore, the following classification levels will only focus on defective subwindows. There are 2094 defective and 22792 non-defective subwindows (hence samples). We train the classifier with 210 defective and 228 non-defective samples. We tabulate the defect detection performances (on test samples only) in Table~\ref{table:defect}.

\begin{table}[htbp]
\caption{Defect detection performances in percentage.} \label{table:defect} \centering
\begin{tabular}{lccc}
\hline
& \multicolumn{3}{c} {Performance ($\%$) } \\ \cline{2-4}
Feature set& defective &non-defective & average\\
\hline
Statistical & 80.0 & 60.2 & 61.7\\
LGSR based& 83.3 & 82.2 & 82.3\\
GMSR based & 87.0 & 85.1 & 85.2\\
\textbf{Combined} & \textbf{89.1} & \textbf{85.4} & \textbf{85.7}\\
\hline
\end{tabular}
\end{table}

Before analyzing the results in Table~\ref{table:defect}, it is better to remaind the format for features. Each GMSR and LGSR based feature has a superscript representing the support region type it is extracted from. Also, each feature has a subscript within parenthesis denoting the color band it is extracted from. In Table~\ref{table:defect}, the statistical feature set is formed of ($\sigma_{(g)}$, $p_{0.2,(g)}$); LGSR based feature set is formed of ($d_{n,(v)}^L$, $\sigma_{(v)}^L$, $\mu_{\tilde{m},(v)}^L$); GMSR based feature set is formed of ($\sigma_{(s)}^G$, $p_{0.2,(g)}^G$, $p_{10,(g)}^G$); and the combined feature set is formed of ($p_{0.02,(r)}^L$, $p_{60,(r)}^L$, $\sigma_{(v)}^G$, $p_{0.2,(g)}^G$). These are the best performing subsets obtained from global search for defect detection.

As can be seen in Table~\ref{table:defect}, the statistical feature set performed worst of all in detecting defects. The GMSR based feature set performed better than the LGSR based one. As we ``combine'' GMSR and LGSR based features, we obtain the best performance, $\mathbf{89.1\%}$ detection with a $\mathbf{14.6 \%}$ false alarm rate. If we consider this result more closely, we come up with the following conclusions. Features obtained from GMSR and LGSR extract useful and different information from wood surface images. Therefore, their combination has a higher discrimination power than each one alone. Moreover, in both LGSR and GMSR based features, non-defective region detection performance is slightly lower than the defective region detection performance. This is most probably because of the texture on wood surfaces. Some textured regions may be falsely labeled as defective by both support regions. As for defective region misses, most probably these defective regions resemble a natural texture. Therefore, that region is falsely labeled as non-defective.

\begin{figure}[htbp]
\centering
\includegraphics[width=3 in]{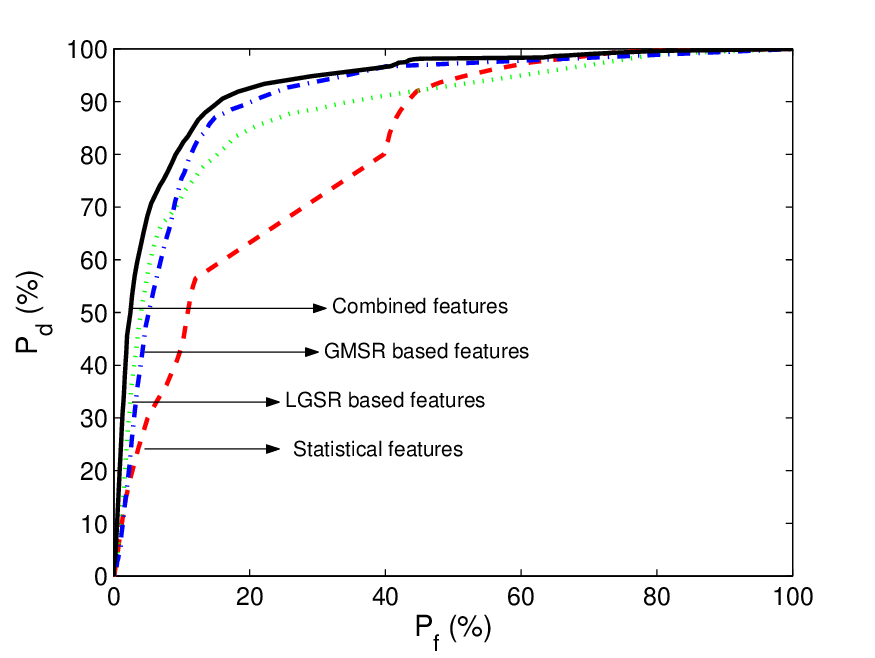}
\caption{ROC curves for defect detection.} \label{fig:ROCdefect}
\end{figure}

Next, we consider the Neyman-Pearson type decision for each feature set tabulated in Table~\ref{table:defect}. We provide the corresponding ROC curves (in percentages) in Fig.~\ref{fig:ROCdefect}. In ROC curves, we desire high $P_d$ for the corresponding low $P_f$ values. Hence, the most desirable ROC curve has the steepest slope in going to the upper left corner of the figure. As can be seen, GMSR and LGSR based features performed better than the statistical features alone according to ROC curves. The combined feature set has the most desirable ROC curve characteristics. This is in accordance with the classification results obtained.

\subsection{Knot and Non-Knot Classification}

Knots are more frequently observed than the remaining defect types. Therefore, they should be considered separately. We next classify defective regions into two classes as knots and non-knots (mostly elongated defect types as resin pocket, core stripe, split, wane, shake, blue stain, brown stain, and bark pocket). There are 1780 knot and 314 non-knot samples at this classification level. We train our Bayes classifier with 36 knot and 63 non-knot samples. We tabulate the classification performances (on test samples only) in Table~\ref{table:knotshake}. In this table, the statistical feature set is formed of ($\sigma_{(r)}$,
$\gamma_{1,(r)}$, $\gamma_{2,(r)}$, $\gamma_{1,(g)}$); LGSR based feature set is formed of ($k_{(r)}^L$, $p_{0.2,(r)}^L$, $p_{90,(r)}^L$, $p_{10,(g)}^L$); GMSR based feature set is formed of ($e_{(r)}^G$, $p_{60,(g)}^G$, $p_{90,(g)}^G$); and the combined feature set is formed of ($\sigma_{(v)}^L$, $\mu_{\tilde{m}(v)}^L$, $\sigma_{(v)}^G$, $p_{60,(r)}^G$, $p_{90,(g)}^G$). These are the best performing subsets obtained from global search for knot and non-knot classification on defective samples.

\begin{table}[htbp]
\centering \caption{Knot and non-knot classification performances in percentage.} \label{table:knotshake}
\begin{tabular}{lccc}
\hline
& \multicolumn{3}{c} {Performance ($\%$) } \\ \cline{2-4}
Feature set& knot & non-knot & average\\
\hline
Statistical & 64.2 & 64.1 & 64.2\\
LGSR based & 84.4 & 81.3 & 84.0\\
\textbf{GMSR based} & \textbf{87.1} & \textbf{85.7} & \textbf{86.9}\\
Combined & 86.1 & 82.1 & 85.6\\
\hline
\end{tabular}
\end{table}

As can be seen in Table~\ref{table:knotshake}, GMSR and LGSR based features performed better than statistical features alone in classification. Furthermore, GMSR based features has the best performance, $\mathbf{87.1 \%}$ correct knot detection with $\mathbf{13.1 \%}$ false alarm rate. The combined feature space could not perform as good as the GMSR based feature space alone in discriminating knots and non-knots. However, the difference between both feature sets is statistically insignificant. The major difference is in detecting non-knots. Here, the GMSR based feature space has a significant advantage over the combined feature set. One may
conclude that, LGSR based features negatively affect the classification performance in the combined feature set. This is most probably because of the support region formation process.

\begin{figure}[htbp]
\centering
\includegraphics[width=3 in]{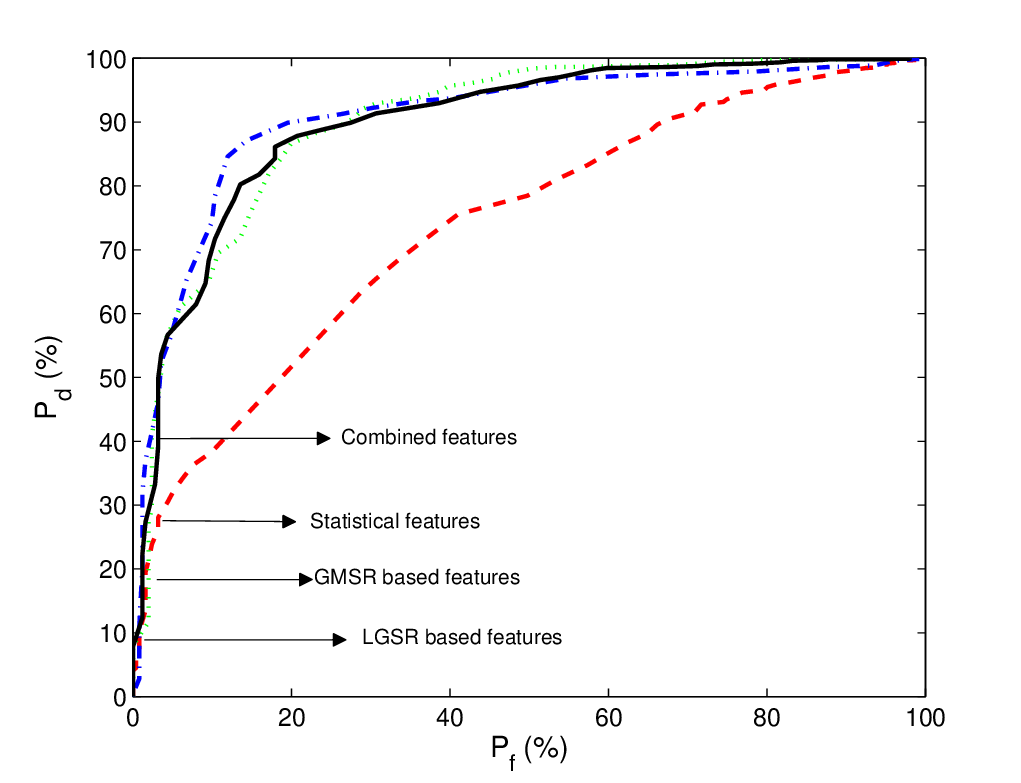}
\caption{ROC curves for the knot and non-knot classification.} \label{fig:ROCshakeknot}
\end{figure}

Similar to the initial defect detection level, we provide the ROC curves for the knot and non-knot discrimination in Fig.~\ref{fig:ROCshakeknot}. The performance of the GMSR based features is also better than other feature spaces in this setup.

\subsection{Dry and Sound Knot Classification}

Finally, we consider discriminating dry and sound knots. These knot types have similar shape characteristics. Sound knots have vague boundaries and are harder to detect. On the other hand, dry knots have sharp boundaries and are easier to detect. Therefore, this final test gives hints about the behavior of our features on discriminating similar shaped defects with different boundary characteristics.

At this classification level, we have 1780 knot samples of which 1508 are dry and 272 are sound knots. We used 33 sound and 356 dry knot samples for training purposes. We next tabulate the discrimination performances for each feature set (on test samples only) in Table~\ref{table:sounddry}. In this table, the statistical feature set is formed of ($p_{10,(r)}$, $\mu_{(g)}$, $\sigma_{(g)}$, $p_{0.2,(g)}$); LGSR based feature set is formed of ($k_{(r)}^L$, $p_{90,(r)}^L$, $p_{60,(g)}^L$); GMSR based feature set is formed of ($p_{0.2,(r)}^G$, $p_{0.02,(g)}^G$); and the combined feature set is formed of ($\sigma_{(v)}^L$, $p_{0.02,(r)}^L$, $p_{10,(g)}^L$, $p_{0.2,(g)}^G$). These are the best performing subsets obtained from global search for dry and sound knot classification.

\begin{table}[htbp]
\caption{Dry and sound knot classification performances in percentage.} \label{table:sounddry} 
\centering
\begin{tabular}{lccc}
\hline
 & \multicolumn{3}{c} {Performance ($\%$) } \\ \cline{2-4}
Feature set& dry knot &sound knot & average\\
\hline
Statistical &  56.4 & 73.7 & 59.3\\
\textbf{LGSR based } &  \textbf{74.5} & \textbf{70.2} & \textbf{73.8}\\
GMSR based &  69.1 & 69.6 & 69.2\\
Combined & 66.1 & 68.2 & 66.4\\
\hline
\end{tabular}
\end{table}

As can be seen in Table~\ref{table:sounddry}, the LGSR based feature set performed best with a $\mathbf{74.5\%}$ dry knot and a $\mathbf{70.2\%}$ sound knot correct classification performance. Since sound knots are harder to detect than dry knots, we obtain a relatively lower classification performance for them. The LGSR based feature set performed better than the rest in detecting dry knots. One possible explanation for this result is that, dry knots have higher contrast around their boundaries. This is detected by the LoG filter better than the gradient magnitude. This may lead to a better performance for the LGSR based feature set. Another important result observed in Table~\ref{table:sounddry} is that, the statistical feature set has a high sound knot detection performance. Since the structure information may not be captured reliably, GMSR and LGSR based features would not perform better than the statistical feature set alone in this classification level.

\begin{figure}[htbp]
\centering
\includegraphics[width=3 in]{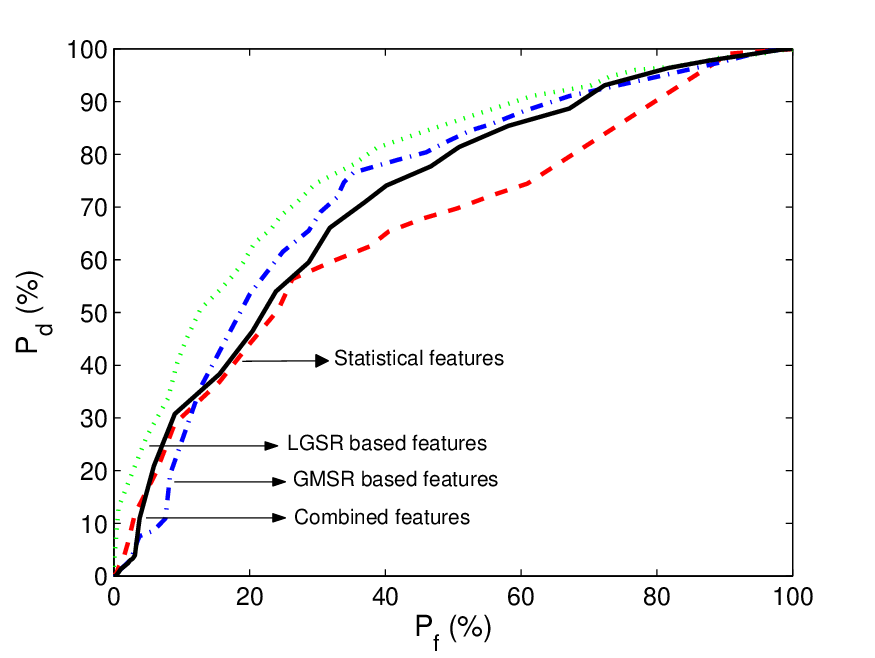}
\caption{ROC curves for dry and sound knot classification.} \label{fig:ROCsounddryknot}
\end{figure}

We provide the corresponding ROC curves in Fig.~\ref{fig:ROCsounddryknot}. As can be seen in this figure, the LGSR based feature set performed better than the rest. The GMSR based feature set has the worst performance around the $10\%$ false alarm rate. However, as the false alarm rate increases, this performance increases beyond the statistical feature set. This is consistent with the previous results tabulated in Table~\ref{table:sounddry}.

\section{Final Comments}\label{sec:final}

Automatically detecting and grading defects on wood surfaces is an important problem. Therefore, this study focuses on defect detection and classification on wood surface images. We introduced novel structural and conditional statistical features for this purpose. We obtain these features from support regions based on the gradient magnitude and LoG filtering (introduced in this study). These two support region extraction methods have specific responses to different defect types on wood surfaces. As we extracted features, we used Bayes classifiers for defect detection and classification. The first level in classification is in detecting defects. Over 24448 test samples, we obtain an average of $\mathbf{85.7\%}$ defect detection performance. The next level in classification is classifying knots and non-knots. We obtain an $\mathbf{86.9\%}$ performance at this level over 1995 test samples. Finally, we classify sound and dry knot samples. At this classification level, we obtain a $\mathbf{73.8\%}$ performance over 1391 test samples. We conclude that our structural and conditional statistical features outperform the existing statistical features in every classification level. The main reason for this performance improvement is adding an extra spatial information (via support regions) to feature calculations. One more advantage of our feature extraction methods is that, we can also obtain the exact position and shape of the defect.

\appendix

\section{Appendix}

In previous sections, we picked the Bayes classifier without any justification. Similarly, we adjusted all parameters to maximize the classification performance. Although some of the parameters are adjusted based on physical reasoning (such as knot size), some of them are obtained experimentally. Here, we justify our classifier and parameter value selections. Therefore, we pick the GMSR based feature space in classifying the defective and non-defective samples as a benchmark.

\subsection{The Effect of the Classifier}

We start with testing the effect of choosing different classifiers in defect detection. Therefore, we picked the statistical Bayes and KNN (with K=5) classifiers. We also used a three layer feedforward neural network (FF-NN) and a learning vector quantization machine (LVQ) both with 8 hidden neurons as classifiers. We tabulate the performance of these four classifiers in Table~\ref{table:classifiers}.

\begin{table}[htbp]
\caption{Performance of different classifiers.} \label{table:classifiers}
\centering
\begin{tabular}{cccc}
\hline
&  \multicolumn{3}{c} {Performance ($\%$) } \\ \cline{2-4}
Classifier  &  defective &non-defective & average\\
\hline
\textbf{Bayes} &  \textbf{87.0} & \textbf{85.1} & \textbf{85.2}\\
KNN &  90.7 & 78.4 & 79.5\\
LVQ &  87.7 & 76.4 & 77.3\\
FF-NN &  88.5 & 83.0 & 83.4\\
\hline
\end{tabular}
\end{table}

As can be seen in Table~\ref{table:classifiers}, the Bayes classifier performed best of all in average classification. Although the KNN classifier has a $90.7 \%$ defect detection performance, it has undesirably low non-defective region detection performance. This is also the case for the LVQ based classifier. Besides, the Bayes classifier is fastest of all. Therefore, using the Bayes classifier seems to be a proper choice in this study.

\subsection{The Effect of the Support Region Size}

In calculating GMSR and LGSR based features, we discard ``small'' support regions. In other saying, we eliminate support regions having areas smaller than a minimum object size. Next, we test the effect of this minimum object size on the classification performance. Therefore, we tested three different minimum object sizes as 10, 50, and 100 pixels. We tabulate the classification performances for each object size in Table~\ref{table:objsize}. As can be seen in this table, the best performance is obtained for the minimum object size of 50 pixels. For other values, the classification performance decreases either in detecting defective samples or in detecting non-defective samples.

\begin{table}[htbp]
\caption{Classification performance wrt. different object size.} \label{table:objsize}
\centering
\begin{tabular}{cccc}
\hline
&  \multicolumn{3}{c} {Performance ($\%$) } \\ \cline{2-4}
Object size  &  defective &non-defective & average\\
\hline
10 &  99.7 & 26.3 & 31.9\\
\textbf{50} &  \textbf{87.0} & \textbf{85.1} & \textbf{85.2}\\
100 &  80.2 & 90.7 & 89.8\\
\hline
\end{tabular}
\end{table}

\subsection{The Effect of the Filter Scale ($\tau_g$)}

Finally, we test the effect of filter scale, $\tau_g$, in gradient calculations. This scale directly affects the support region formation. A low $\tau_g$ compared to the wood surface texture size will lead to many false alarms. Similarly, a high $\tau_g$ will lead to missing most defects. We pick three $\tau_g$ values and tabulate the corresponding classification performances in Table~\ref{table:filtersize}. As can be seen in this table, $\tau_g=2$ seems to be the most appropriate choice.

\begin{table}[htbp]
\caption{Classification performance wrt. different $\tau_g$ in gradient calculations.} \label{table:filtersize} \centering
\begin{tabular}{cccc}
\hline
&  \multicolumn{3}{c} {Performance ($\%$) } \\ \cline{2-4}
$\tau_g$  &  defective &non-defective & average\\
\hline
1 &  86.8 & 85.2 & 85.3\\
\textbf{2} &  \textbf{87.0} & \textbf{85.1} & \textbf{85.2}\\
3 &  86.8 & 85.3 & 85.4\\
\hline
\end{tabular}
\end{table}

\bibliographystyle{elsarticle-harv}
\bibliography{woodinspection}

\end{document}